\pdfoutput=1
\documentclass{article} % For LaTeX2e
\usepackage{iclr2020_conference,times}

\usepackage{amsmath,amsfonts,bm}

\def\eqref#1{equation~\ref{#1}}
\def\1{\bm{1}}

\DeclareMathAlphabet{\mathsfit}{\encodingdefault}{\sfdefault}{m}{sl}
\SetMathAlphabet{\mathsfit}{bold}{\encodingdefault}{\sfdefault}{bx}{n}

\usepackage{hyperref}
\usepackage{url}
\usepackage{graphicx}
\usepackage{algorithm}
\usepackage{algorithmic}
\usepackage{multicol}
\usepackage{multirow}

\title{Goal-conditioned Batch Reinforcement\\Learning for Rotation Invariant Locomotion}

\author{Aditi Mavalankar \\
University of California, San Diego\\
\texttt{amavalan@eng.ucsd.edu} \\
}

\iclrfinalcopy
\begin{document}

\maketitle

\vspace{-0.5cm}
\begin{abstract}
We propose a novel approach to learn goal-conditioned policies for locomotion in a batch RL setting. The batch data is collected by a policy that is \textit{not} goal-conditioned. For the locomotion task, this translates to data collection using a policy learnt by the agent for walking straight in one direction, and using that data to learn a goal-conditioned policy that enables the agent to walk in any direction. The data collection policy used should be invariant to the direction the agent is facing i.e.\ regardless of its initial orientation, the agent should take the same actions to walk forward. We exploit this property to learn a goal-conditioned policy using two key ideas: (1) augmenting data by generating trajectories with the same actions in different directions, and (2) learning an encoder that enforces invariance between these rotated trajectories with a Siamese framework. We show that our approach outperforms existing RL algorithms on 3-D locomotion agents like Ant, Humanoid and Minitaur.
\end{abstract}

\vspace{-0.2cm}
\section{Introduction}

Goal-conditioned reinforcement learning (RL) algorithms (\cite{kaelbling1993learning, Schaul:2015:UVF:3045118.3045258}) learn policies that enable the agent to achieve a diverse set of goals. It is challenging to learn such policies due to the goal expanding the dimensionality of the input space. This problem becomes more evident in high-dimensional continuous control tasks like goal-directed locomotion. Since the agent needs to know how to walk in order to walk in different directions, goal-directed locomotion is more challenging than the standard locomotion task i.e.\ walking in a single direction.

We consider the following batch RL (\cite{lange2012batch}) setting to address the goal-directed locomotion task: given a batch consisting of data collected while training a policy for locomotion (similar to the \textit{final buffer} setting in \cite{fujimoto2019off}), we use it to learn a policy for goal-directed locomotion. We refer to this modified setting as \textit{goal-conditioned batch RL}. 

Why do we need this modified setting? This question can be answered based on the following intuitions: (1) learning to walk in one direction is easier than learning to walk in all directions at once, and (2) an agent that can walk in one direction should be able to walk in any direction without a large number of additional environment interactions. These intuitions are specific to the goal-directed locomotion task; however, our approach can be used to address tasks that possess an invariance with respect to their state and/or goal spaces. Here, this structure exists in the form of rotation invariance of the state and goal spaces with respect to the locomotion policy and can be exploited to learn goal-conditioned policies efficiently.

\begin{figure}
    \centering
    \vspace{-1cm}
    \includegraphics[width=0.9\linewidth]{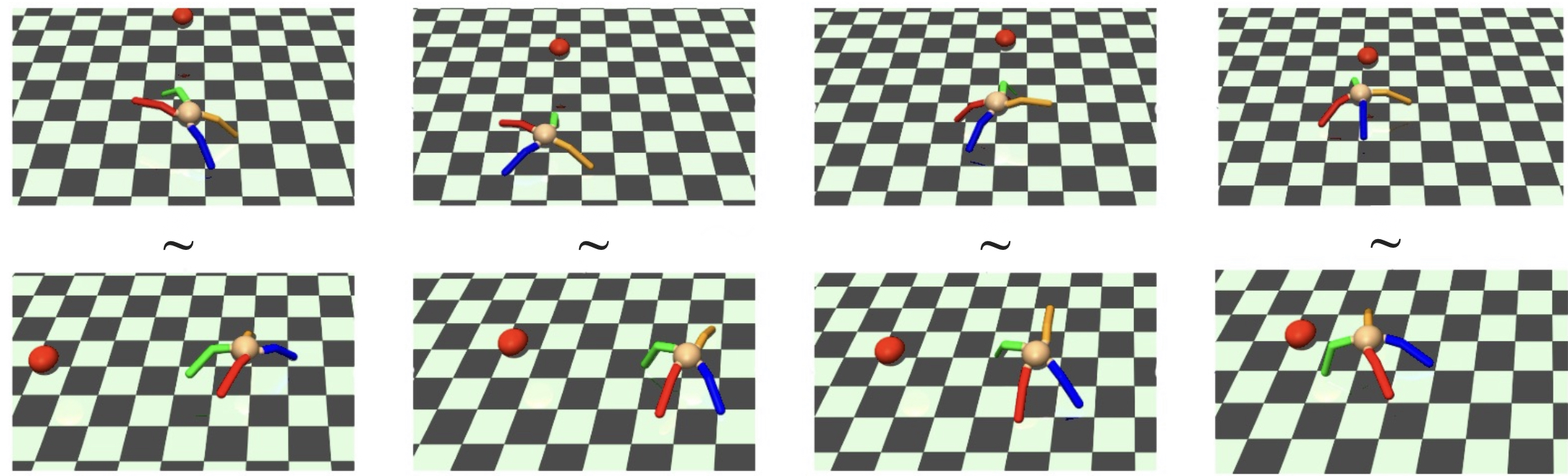}\\
    \hspace{18mm} $t=t_1$ \hfill $t=t_2$ \hfill $t=t_3$ \hfill $t=t_4$ \hspace{17mm}
    \caption{An example of \textit{equivalent} trajectories. The first row shows intermediate steps in trajectory $\tau$, and the second row shows the corresponding intermediate steps in $\tilde{\tau}$. In both trajectories, the agent takes the same actions to reach the goal represented by the red sphere.}
    \label{fig:equivalence}
    \vspace{-0.1cm}
\end{figure}

Our approach exploits the intuitions listed above making use of two well-known techniques: (1) data augmentation and (2) representation learning. Data augmentation involves simulating the agent's actions observed in the batch trajectories in different directions resulting in an augmented batch containing trajectories equivalent to those observed in the original batch (see Fig. \ref{fig:equivalence}). We then use representation learning to learn embeddings that capture this equivalence between trajectories in the augmented batch. We show the results of our proposed approach on the Ant environment in OpenAI Gym (\cite{gym, mujoco}), and the Humanoid and Minitaur environments in Pybullet (\cite{pybullet}). On all three environments, our method outperforms existing methods using standard RL techniques, as well as goal-conditioned RL techniques. We also compare our approach to baselines in the goal-conditioned batch RL setting. Here, the batch consists of either random samples, or on-policy samples collected by the agent. Data augmentation alone beats all the above baselines, and using representation learning further boosts performance.

\section{Goal-conditioned Batch RL}
\label{goal-conditioned-batch-rl}

\begin{figure}
    \vspace{-1cm}
    \begin{center}
    \includegraphics[width=0.8\linewidth]{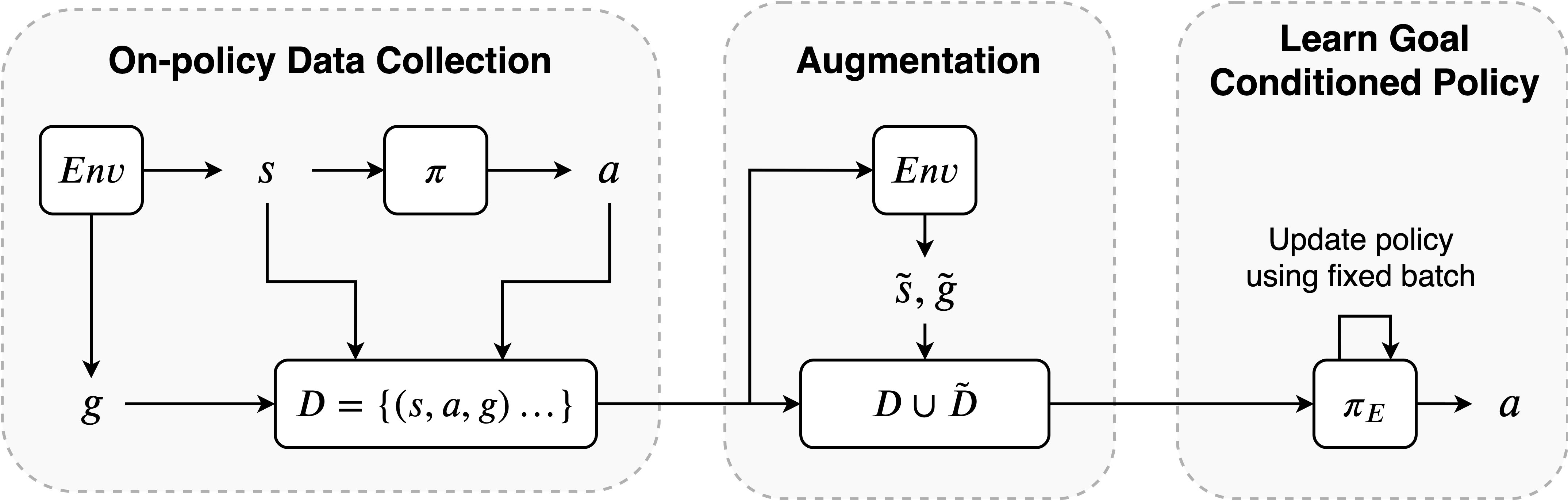}
    \caption{The goal-conditioned batch RL setting. There are 3 stages: (1) Data collection, using any RL algorithm to learn a policy $\pi$ conditioned on the state $s$. All transition data $s$, $a$ and $g$, are stored in $\mathcal{D}$. (2) Data augmentation, where $\mathcal{D}$ can be augmented by utilizing the rotation invariance of $\mathcal{S}$ and $\mathcal{G}$, resulting in $\tilde{\mathcal{D}}$, and (3) Learning the goal-conditioned policy $\pi_E$ from the batch data $\mathcal{D} \cup \tilde{\mathcal{D}}$.}
    \label{fig:goal-conditioned-batch-rl}
    \end{center}
    %\vspace{-0.4cm}
\end{figure}

In the goal-conditioned batch RL setting, the aim is to learn a goal-conditioned policy from a batch of data collected by a policy that is \textit{not} goal-conditioned. This setting is shown in Fig. \ref{fig:goal-conditioned-batch-rl}. We describe data collection and the two key aspects of our algorithm - data augmentation and learning equivalence between trajectories in the augmented dataset. We also discuss an approach to learn a na\"ive goal-conditioned policy from the data as a baseline, in \ref{learning-gcp}.

\textbf{Data collection.} We collect batch data $\mathcal{D}$ using a standard RL algorithm (we use Proximal Policy Optimization (PPO) (\cite{ppo}) and Soft Actor-Critic (SAC) (\cite{pmlr-v80-haarnoja18b})) to train a policy $\pi: \mathcal{S} \rightarrow \mathcal{A}$. $\mathcal{D} = \{(s, a, g), \dots\}$, where $s$ is the state, $a$ is the action, and $g$ is the one-step goal. Note that $\pi$ is not conditioned on $g$. Here, $g$ is the 3-D position of the agent.

\textbf{Equivalence.} We emphasize the property that an agent's policy for locomotion should be invariant to the direction in which locomotion occurs i.e.\ an agent's policy for walking straight ahead should be the same, regardless of which direction it is facing. We use this to augment $\mathcal{D}$ with trajectories that are \textit{equivalent} to those present in it. $\tau_1 = \{s_1, a_1, g_1, s_2, a_2, g_2, ..., s_T, a_T, g_T\}$ and $\tau_2 = \{\tilde{s}_1, \tilde{a}_1, \tilde{g}_1, \tilde{s}_2, \tilde{a}_2, \tilde{g}_2, \dots, \tilde{s}_T, \tilde{a}_T, \tilde{g}_T\}$ are equivalent trajectories if for any $\theta \in [0, 2\pi), \tilde{a}_i = a_i, \tilde{s}_i = rot(s_i, \theta), \tilde{g}_1 = rot(g_1, \theta), \tilde{g}_2 = rot(g_2, \theta), ..., \tilde{g}_T = rot(g_T, \theta)$, for $i = \{1, ..., T\}$ where $rot(x, \theta)$ is a rotation of $x$ about the agent by angle $\theta$. Fig. \ref{fig:equivalence} shows an example of such trajectories.

\textbf{Data augmentation.} We use the above definition of equivalence to augment $\mathcal{D}$. For each trajectory $\tau$ observed in the dataset $\mathcal{D}$, we generate another trajectory $\tilde{\tau}$ that is equivalent to it by randomly choosing $\theta \in [0, 2\pi)$. The augmented dataset $\mathcal{D}_{aug}$ consists of all trajectories in $\mathcal{D}$ and the generated set of their equivalent trajectories $\tilde{\mathcal{D}}$. $\tilde{\tau}$ is collected by rotating the initial configuration of the agent in $\tau$, and then repeating the actions taken by the agent in $\tau$ to expand the batch data. This concept is illustrated in Fig. \ref{fig:equivalence} (algorithm in \ref{data-aug}).

\textbf{Our Approach: Enforcing equivalence.} Once we have the augmented dataset $\tilde{\mathcal{D}}$, we focus on learning a model that incorporates this equivalence into the training procedure. To do this, we learn two models: an encoder $E$, and a policy $\pi_{E}$. The encoder is defined as $E \colon \mathcal{S} \times \mathcal{G} \rightarrow \mathbb{R}^k$. It outputs $h = E(s, g)$, where $h$ is a $k$-dimensional embedding produced by the encoder. The goal-conditioned policy with equivalence is defined as $\pi_E \colon \mathbb{R}^k \rightarrow \mathcal{A}, \pi_E(h) = a$. $h$ should capture the equivalence between $(s, g)$ and $(\tilde{s}, \tilde{g})$. Thus, if $(s, g) \sim (\tilde{s}, \tilde{g})$, $h = \tilde{h}$, where $h = E(s, g)$ and $\tilde{h} = E(\tilde{s}, \tilde{g})$. We generate these embeddings for the equivalent samples simultaneously; thus, we use a Siamese framework (\cite{koch2015siamese}) to learn $E$ i.e.\ $h$ and $\tilde{h}$ are computed using shared weights. Both $E$ and $\pi_E$ are MLPs, learnt using the objective functions $\mathcal{L}_{enc}$ and $\mathcal{L}_{\pi_E}$ respectively. 

\vspace{-0.6cm}
\begin{align*}
     \mathcal{L}_{enc} = \frac{1}{M}\sum_{i=1}^M||h_i - \tilde{h}_i||_2^2   &   &   \mathcal{L}_{\pi_E} = \frac{1}{M}\sum_{i=1}^M||\pi_E(\bar h_i) - a_i||_2^2\\
\end{align*}
\vspace{-1cm}

where $h_i = E(s_i, g_i)$, $\tilde{h}_i = E(\tilde{s}_i, \tilde{g}_i)$, and $\bar{h}_i = \frac{1}{2}(h_i + \tilde{h}_i)$. The overall loss is given by $\mathcal{L} = \lambda \mathcal{L}_{enc} + (1 - \lambda) \mathcal{L}_{\pi_E}$, where $\lambda \in [0, 1]$. The procedure is described in Algorithm \ref{alg:our-approach}.

\section{Experiments}
\label{experiments}
\vspace{-0.2cm}

We perform experiments on the goal-directed locomotion task for bipedal and quadrupedal agents: Ant, Humanoid and Minitaur.

\begin{algorithm}
  \caption{Enforcing equivalence}
  \label{alg:our-approach}
\begin{algorithmic}
  \STATE Given $\mathcal{D}_{aug} = \{\tau_1, \tilde{\tau}_1, \tau_2, \tilde{\tau}_2, ..., \tau_N, \tilde{\tau}_N\}$, learning rate $\eta$, encoder $E$ with parameters $\phi_{enc}$, goal-conditioned policy with equivalence $\pi_E$ with parameters $\phi_{\pi_E}$
  \STATE Initialize $\phi_{enc}$ and $\phi_{\pi_E}$ randomly
  \REPEAT
      \FOR{$(\tau, \tilde{\tau}) \in \mathcal{D}_{aug}$}
        \STATE $s_1, a_1, g_1, s_2, a_2, g_2, ..., s_T, a_T, g_T \leftarrow \tau$
        \STATE $\tilde{s}_1, a_1, \tilde{g}_1, \tilde{s}_2, a_2, \tilde{g}_2, ..., \tilde{s}_T, a_T, \tilde{g}_T \leftarrow \tilde{\tau}$
        \FOR{$t=1$ {\bfseries to} $T$}
            \STATE $h_t = E(s_t, g_t)$
            \STATE $\tilde{h}_t = E(\tilde{s}_t, \tilde{g}_t)$
            \STATE $\bar{h}_t = \frac{1}{2}(h_t + \tilde{h}_t)$
        \ENDFOR
        \STATE Compute $\mathcal{L} = \lambda \mathcal{L}_{enc} + (1 - \lambda) \mathcal{L}_{\pi_E}$
        \STATE $\phi_{enc} \leftarrow \phi_{enc} - \eta \nabla \mathcal{L}$
        \STATE $\phi_{\pi_E} \leftarrow \phi_{\pi_E} - \eta \nabla \mathcal{L}$
      \ENDFOR
  \UNTIL{$n$ iterations}
\end{algorithmic}
\end{algorithm}

\textbf{Standard RL}. We use a standard on-policy or off-policy RL algorithm to train a goal-conditioned policy. We use SAC for Humanoid, and PPO for Ant and Minitaur.

\textbf{Goal-conditioned RL}. We use Hindsight Experience Replay (HER) (\cite{andrychowicz2017hindsight}) to train the goal-conditioned policy with SAC and Deep Deterministic Policy Gradient (DDPG) (\cite{ddpg}) as the off-policy RL algorithms. We report results on the method with the best performance on the goal-directed locomotion task on both dense and sparse reward environments. 

\textbf{Goal-conditioned batch RL}. We discuss different variants of the goal-conditioned batch RL setting described in Section \ref{goal-conditioned-batch-rl}. In each case except random samples, the batch consists of data collected while training SAC for Humanoid, and PPO for Ant and Minitaur. \textbf{(1) Random samples}: We learn a na\"ive goal-conditioned policy $\pi_g$ over a batch consisting of random samples collected from the environment. \textbf{(2) On-policy samples}: We learn $\pi_g$ over a batch containing data collected while training the agent for locomotion in one direction. \textbf{(3) Augmented samples}: We use data augmentation to collect samples that are \textit{equivalent} to the on-policy samples and learn $\pi_g$. \textbf{(4) Equivalence on augmented samples}: We follow the procedure in Alg. \ref{alg:our-approach} to learn an encoder $E$ and goal-conditioned policy with equivalence $\pi_E$.

\textbf{Implementation details}. 
All 3 models - $\pi_g$,  $E$ and $\pi_E$, are Multilayer Perceptrons (MLPs) with 2 hidden layers. Exact details of model architectures for each experiment, number of samples of environment interaction, optimizers, environments, reward functions, and other experimental details are provided in the Appendix.

\begin{figure}
  \vspace{-0.4cm}
  \centering
  \includegraphics[width=0.329\linewidth]{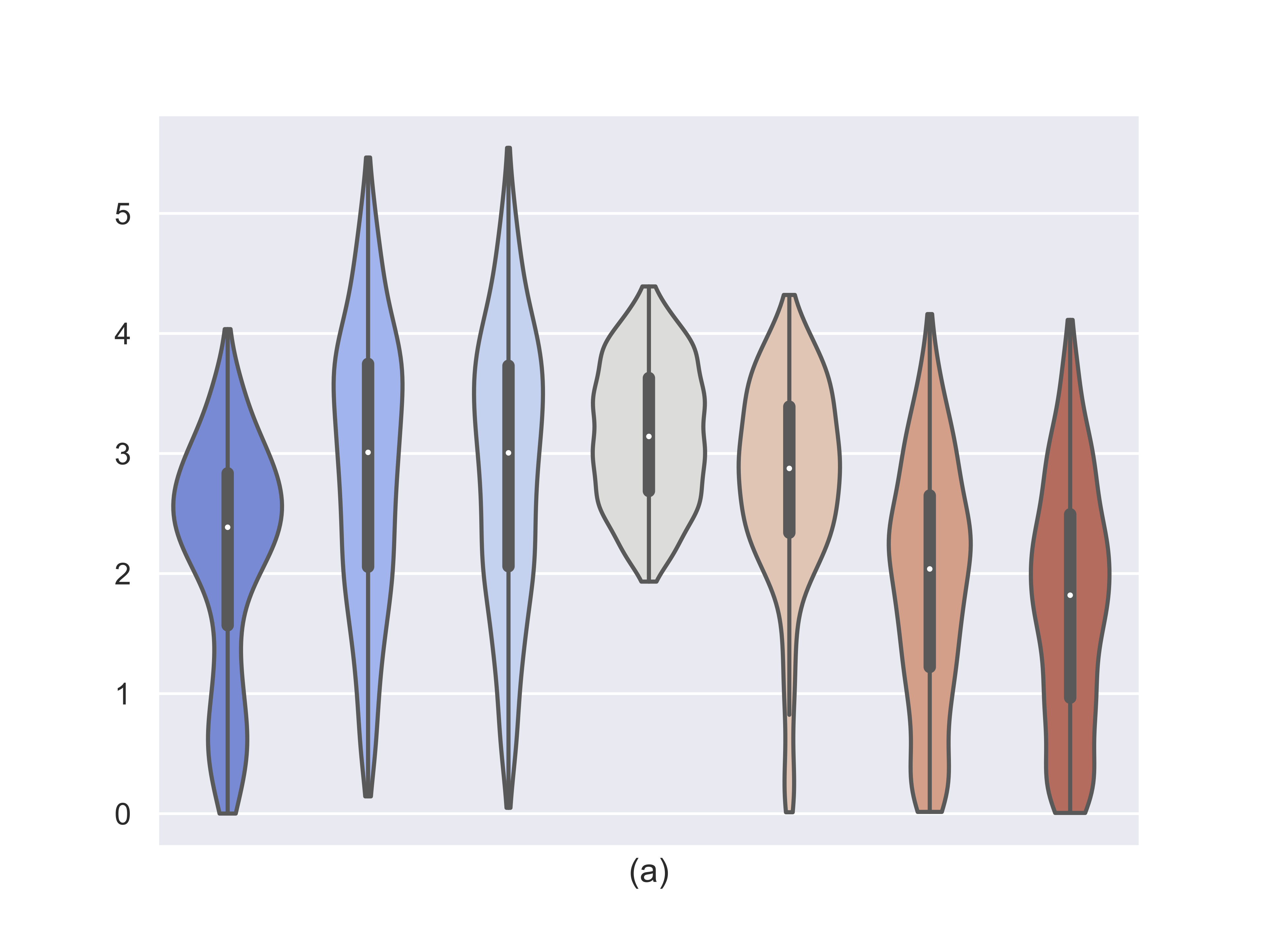}
  \includegraphics[width=0.329\linewidth]{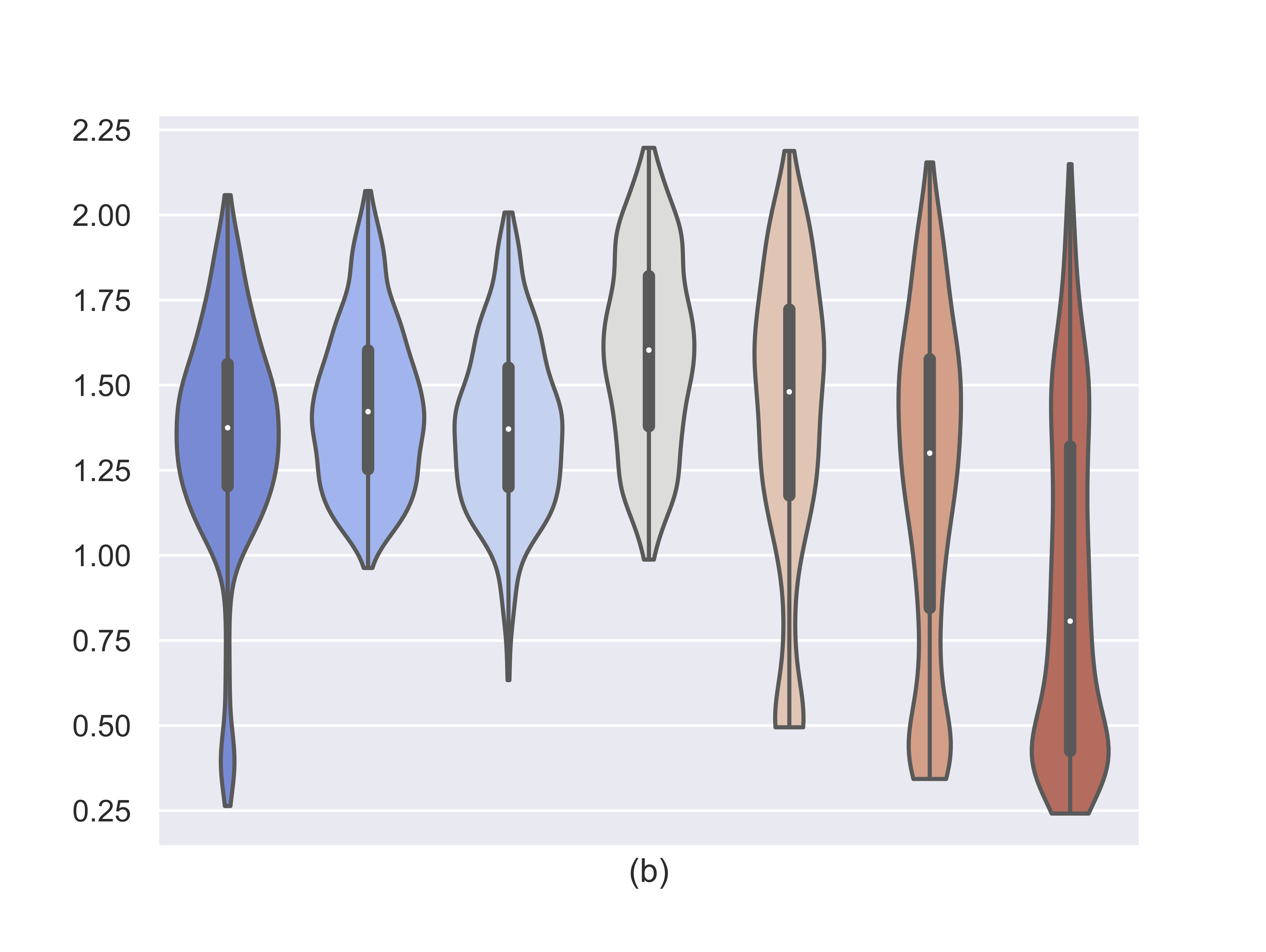}
  \includegraphics[width=0.329\linewidth]{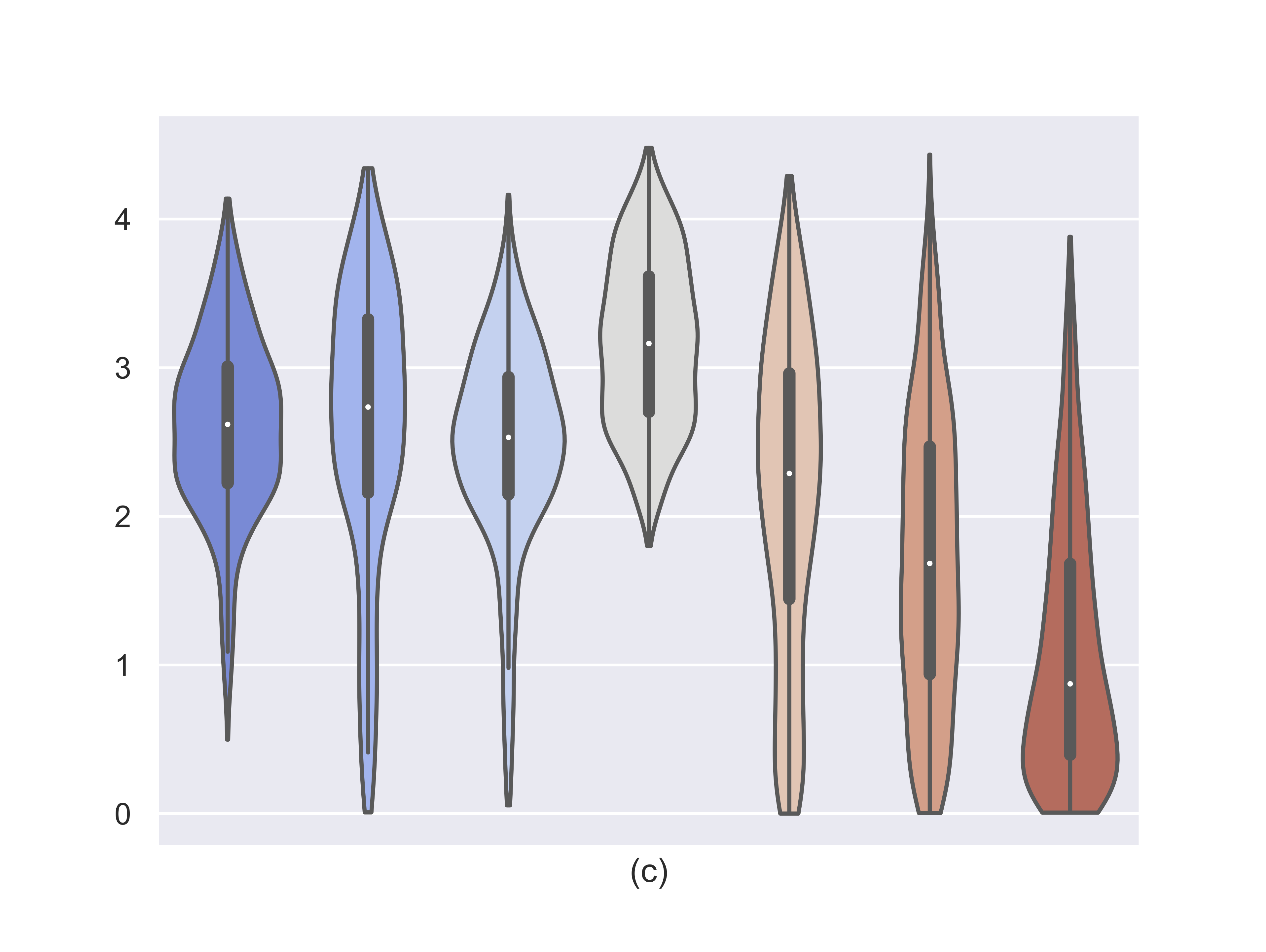}\\
  \includegraphics[width=0.95\linewidth]{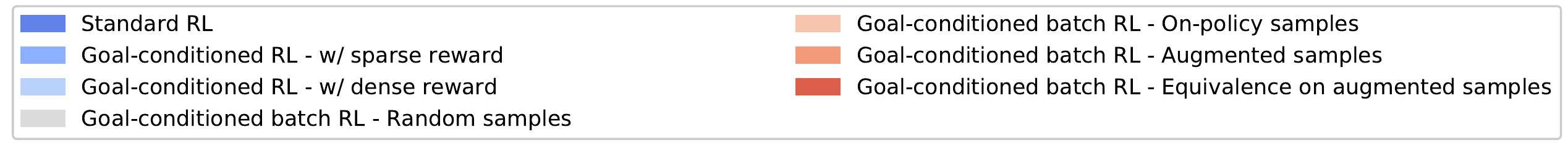}
  \caption{Violin plots showing the distribution of closest distance from the goal for each episode for the 3 environments: (a) Humanoid (b) Minitaur (c) Ant.}
  \label{fig:violin_plots}
  \vspace{-0.5cm}
\end{figure}

\textbf{Results}. We compare the performances of these experiments based on the closest distance to the goal that the agent is able to achieve. We report the mean and standard deviation of the closest distance to the goal for each agent, averaged over 100 runs on 10 random seeds. We show the results of our approach, compared with all the other baselines discussed above, in Table \ref{tab:results} and Fig. \ref{fig:violin_plots}. In each environment, our approach outperforms all baselines considered. The second-best approach across all environments is learning a na\"ive goal-conditioned policy on augmented samples. This indicates the effectiveness of both our key contributions: data augmentation and enforcing equivalence. Learning a na\"ive goal-conditioned policy on augmented samples improves performance over all methods because the actions taken are consistent with the definition of equivalence. Enforcing equivalence using the Siamese framework forces the same embedding to be learnt for all equivalent state-goal configurations, thus generalizing to a wider set of goals than all the other baselines.

\begin{table}
    \vspace{-1cm}
    \hspace{-1cm}
    \begin{tabular}{lllllll}
        \hline
        \textbf{Model} & \multicolumn{2}{l}{\textbf{Humanoid}} &\multicolumn{2}{l}{\textbf{Minitaur}} &\multicolumn{2}{l}{\textbf{Ant}}\\
         & Mean & Std. dev. & Mean & Std. dev. & Mean & Std. dev.\\
        \hline
        \textit{Standard RL} & & \\
        \cite{ppo, pmlr-v80-haarnoja18b} & 2.160 & 0.955 & 1.358 & 0.326 & 2.608 & 0.607\\
        \hline
        \textit{Goal-conditioned RL} \cite{andrychowicz2017hindsight} & & \\
        w/ sparse reward & 2.902 & 1.138 & 1.440 & 0.237 & 2.602 & 0.953\\
        w/ dense reward & 2.891 & 1.135 & 1.380 & 0.251 & 2.475 & 0.693\\
        \hline
        \textit{Goal-conditioned batch RL} & & \\
        Random samples & 3.159 & 0.579 & 1.598 & 0.282 & 3.166 & 0.569\\
        On-policy samples & 2.777 & 0.841 & 1.403 & 0.438 & 2.155 & 1.054\\
        Augmented samples & 1.936 & 0.977 & 1.212 & 0.485 & 1.721 & 0.978\\
        \textbf{Our approach: Enforcing Equivalence} & \textbf{1.779} & \textbf{0.986} & \textbf{0.904} & \textbf{0.482} & \textbf{1.105} & \textbf{0.866}\\
        \hline
    \end{tabular}
    \caption{Results of our algorithm (in bold) compared with all the baselines discussed in Section \ref{experiments}.}
    \label{tab:results}
\end{table}

\vspace{-0.3cm}
\section{Related Work}
\label{related-work}
\vspace{-0.3cm}

Goal-conditioned RL has been used in hierarchical RL algorithms (\cite{NIPS2016_6233, NIPS2018_7591}), as a link between model-based and model-free RL for control (\cite{pong*2018temporal}), with the imitation learning framework (\cite{ding2019goal}), as a pretrained prior to learn actionable aspects of the state and goal spaces (\cite{DBLP:journals/corr/abs-1811-07819}), and to perform higher-level tasks like planning (\cite{nasiriany2019planning}). Similar to our approach that uses equivalence between trajectories, some recent works (\cite{mishra2019augmenting, abdolhosseini2019learning}) exploit symmetry, albeit in the agent than in the environment, to learn policies that result in better gait during locomotion. A number of batch RL methods have been proposed recently, addressing large distributional shift in continuous control (\cite{fujimoto2019off}), learning under constraints (\cite{le2019batch}), etc. Our approach extends the batch RL formulation to goal-conditioned RL and exploits symmetry in the state and goal spaces to learn efficient goal-conditioned policies for goal-directed locomotion.

\vspace{-0.3cm}
\section{Conclusion}
\label{conclusion}
\vspace{-0.3cm}

We propose a modified batch RL setting, termed goal-conditioned batch RL, to solve the goal-directed locomotion task for high-dimensional continuous control agents. Our method utilizes invariance to rotation and combines data augmentation with representation learning, resulting in agents being able to reach different goals more successfully than those trained using existing RL and goal-conditioned RL methods.

\section{Acknowledgements}
Thanks to Abhinav Moudgil, Jonathan J Hunt, Quan Vuong, and Sicun Gao for insightful discussions and helpful feedback on the topic and paper, and to Sophia Sun, Gary Cottrell, Zhi Wang, Lawrence Saul, Geelon So, Vraj Shah, and Nishant Bhaskar for their detailed reviews of the paper.

\bibliography{iclr2020_conference}

\begin{thebibliography}{24}
\providecommand{\natexlab}[1]{#1}
\providecommand{\url}[1]{\texttt{#1}}
\expandafter\ifx\csname urlstyle\endcsname\relax
  \providecommand{\doi}[1]{doi: #1}\else
  \providecommand{\doi}{doi: \begingroup \urlstyle{rm}\Url}\fi

\bibitem[Abdolhosseini et~al.(2019)Abdolhosseini, Ling, Xie, Peng, and van~de
  Panne]{abdolhosseini2019learning}
Farzad Abdolhosseini, Hung~Yu Ling, Zhaoming Xie, Xue~Bin Peng, and Michiel
  van~de Panne.
\newblock On learning symmetric locomotion.
\newblock In \emph{Motion, Interaction and Games}, pp.\  1--10. 2019.

\bibitem[Andrychowicz et~al.(2017)Andrychowicz, Wolski, Ray, Schneider, Fong,
  Welinder, McGrew, Tobin, Abbeel, and Zaremba]{andrychowicz2017hindsight}
Marcin Andrychowicz, Filip Wolski, Alex Ray, Jonas Schneider, Rachel Fong,
  Peter Welinder, Bob McGrew, Josh Tobin, OpenAI~Pieter Abbeel, and Wojciech
  Zaremba.
\newblock Hindsight experience replay.
\newblock In \emph{Advances in Neural Information Processing Systems}, pp.\
  5048--5058, 2017.

\bibitem[Brockman et~al.(2016)Brockman, Cheung, Pettersson, Schneider,
  Schulman, Tang, and Zaremba]{gym}
Greg Brockman, Vicki Cheung, Ludwig Pettersson, Jonas Schneider, John Schulman,
  Jie Tang, and Wojciech Zaremba.
\newblock Openai gym.
\newblock \emph{CoRR}, abs/1606.01540, 2016.

\bibitem[Coumans \& Bai(2016)Coumans and Bai]{pybullet}
Erwin Coumans and Yunfei Bai.
\newblock Pybullet, a python module for physics simulation for games, robotics
  and machine learning.
\newblock \emph{GitHub repository}, 2016.

\bibitem[Dhariwal et~al.(2017)Dhariwal, Hesse, Klimov, Nichol, Plappert,
  Radford, Schulman, Sidor, Wu, and Zhokhov]{baselines}
Prafulla Dhariwal, Christopher Hesse, Oleg Klimov, Alex Nichol, Matthias
  Plappert, Alec Radford, John Schulman, Szymon Sidor, Yuhuai Wu, and Peter
  Zhokhov.
\newblock Openai baselines.
\newblock \url{https://github.com/openai/baselines}, 2017.

\bibitem[Ding et~al.(2019)Ding, Florensa, Abbeel, and Phielipp]{ding2019goal}
Yiming Ding, Carlos Florensa, Pieter Abbeel, and Mariano Phielipp.
\newblock Goal-conditioned imitation learning.
\newblock In \emph{Advances in Neural Information Processing Systems}, pp.\
  15298--15309, 2019.

\bibitem[Fujimoto et~al.(2019)Fujimoto, Meger, and Precup]{fujimoto2019off}
Scott Fujimoto, David Meger, and Doina Precup.
\newblock Off-policy deep reinforcement learning without exploration.
\newblock In \emph{International Conference on Machine Learning}, pp.\
  2052--2062, 2019.

\bibitem[Ghosh et~al.(2018)Ghosh, Gupta, and
  Levine]{DBLP:journals/corr/abs-1811-07819}
Dibya Ghosh, Abhishek Gupta, and Sergey Levine.
\newblock Learning actionable representations with goal-conditioned policies.
\newblock \emph{CoRR}, abs/1811.07819, 2018.

\bibitem[Haarnoja et~al.(2018)Haarnoja, Zhou, Abbeel, and
  Levine]{pmlr-v80-haarnoja18b}
Tuomas Haarnoja, Aurick Zhou, Pieter Abbeel, and Sergey Levine.
\newblock Soft actor-critic: Off-policy maximum entropy deep reinforcement
  learning with a stochastic actor.
\newblock In \emph{Proceedings of the 35th International Conference on Machine
  Learning}, volume~80 of \emph{Proceedings of Machine Learning Research}, pp.\
   1861--1870, Stockholmsmässan, Stockholm Sweden, 10--15 Jul 2018. PMLR.

\bibitem[Hill et~al.(2018)Hill, Raffin, Ernestus, Gleave, Kanervisto, Traore,
  Dhariwal, Hesse, Klimov, Nichol, Plappert, Radford, Schulman, Sidor, and
  Wu]{stable-baselines}
Ashley Hill, Antonin Raffin, Maximilian Ernestus, Adam Gleave, Anssi
  Kanervisto, Rene Traore, Prafulla Dhariwal, Christopher Hesse, Oleg Klimov,
  Alex Nichol, Matthias Plappert, Alec Radford, John Schulman, Szymon Sidor,
  and Yuhuai Wu.
\newblock Stable baselines.
\newblock \url{https://github.com/hill-a/stable-baselines}, 2018.

\bibitem[Kaelbling(1993)]{kaelbling1993learning}
Leslie~Pack Kaelbling.
\newblock Learning to achieve goals.
\newblock In \emph{Proc. of IJCAI-93}, pp.\  1094--1098, 1993.

\bibitem[Kingma \& Ba(2015)Kingma and Ba]{DBLP:journals/corr/KingmaB14}
Diederik~P. Kingma and Jimmy Ba.
\newblock Adam: {A} method for stochastic optimization.
\newblock In \emph{3rd International Conference on Learning Representations,
  {ICLR} 2015}, 2015.

\bibitem[Koch et~al.(2015)Koch, Zemel, and Salakhutdinov]{koch2015siamese}
Gregory Koch, Richard Zemel, and Ruslan Salakhutdinov.
\newblock Siamese neural networks for one-shot image recognition.
\newblock In \emph{ICML deep learning workshop}, volume~2, 2015.

\bibitem[Kulkarni et~al.(2016)Kulkarni, Narasimhan, Saeedi, and
  Tenenbaum]{NIPS2016_6233}
Tejas~D Kulkarni, Karthik Narasimhan, Ardavan Saeedi, and Josh Tenenbaum.
\newblock Hierarchical deep reinforcement learning: Integrating temporal
  abstraction and intrinsic motivation.
\newblock In \emph{Advances in Neural Information Processing Systems 29}, pp.\
  3675--3683. 2016.

\bibitem[Lange et~al.(2012)Lange, Gabel, and Riedmiller]{lange2012batch}
Sascha Lange, Thomas Gabel, and Martin Riedmiller.
\newblock Batch reinforcement learning.
\newblock In \emph{Reinforcement learning}, pp.\  45--73. Springer, 2012.

\bibitem[Le et~al.(2019)Le, Voloshin, and Yue]{le2019batch}
Hoang Le, Cameron Voloshin, and Yisong Yue.
\newblock Batch policy learning under constraints.
\newblock In \emph{International Conference on Machine Learning}, pp.\
  3703--3712, 2019.

\bibitem[Lillicrap et~al.(2016)Lillicrap, Hunt, Pritzel, Heess, Erez, Tassa,
  Silver, and Wierstra]{ddpg}
Timothy~P Lillicrap, Jonathan~J Hunt, Alexander Pritzel, Nicolas Heess, Tom
  Erez, Yuval Tassa, David Silver, and Daan Wierstra.
\newblock Continuous control with deep reinforcement learning.
\newblock \emph{ICLR 2016}, 2016.

\bibitem[Mishra et~al.(2019)Mishra, Abdolmaleki, Guez, Trochim, and
  Precup]{mishra2019augmenting}
Shruti Mishra, Abbas Abdolmaleki, Arthur Guez, Piotr Trochim, and Doina Precup.
\newblock Augmenting learning using symmetry in a biologically-inspired domain.
\newblock \emph{arXiv preprint arXiv:1910.00528}, 2019.

\bibitem[Nachum et~al.(2018)Nachum, Gu, Lee, and Levine]{NIPS2018_7591}
Ofir Nachum, Shixiang~(Shane) Gu, Honglak Lee, and Sergey Levine.
\newblock Data-efficient hierarchical reinforcement learning.
\newblock In S.~Bengio, H.~Wallach, H.~Larochelle, K.~Grauman, N.~Cesa-Bianchi,
  and R.~Garnett (eds.), \emph{Advances in Neural Information Processing
  Systems 31}, pp.\  3303--3313. 2018.

\bibitem[Pong* et~al.(2018)Pong*, Gu*, Dalal, and Levine]{pong*2018temporal}
Vitchyr Pong*, Shixiang Gu*, Murtaza Dalal, and Sergey Levine.
\newblock Temporal difference models: Model-free deep {RL} for model-based
  control.
\newblock In \emph{International Conference on Learning Representations}, 2018.

\bibitem[Schaul et~al.(2015)Schaul, Horgan, Gregor, and
  Silver]{Schaul:2015:UVF:3045118.3045258}
Tom Schaul, Dan Horgan, Karol Gregor, and David Silver.
\newblock Universal value function approximators.
\newblock In \emph{Proceedings of the 32Nd International Conference on
  International Conference on Machine Learning - Volume 37}, ICML'15, pp.\
  1312--1320. JMLR.org, 2015.

\bibitem[Schulman et~al.(2017)Schulman, Wolski, Dhariwal, Radford, and
  Klimov]{ppo}
John Schulman, Filip Wolski, Prafulla Dhariwal, Alec Radford, and Oleg Klimov.
\newblock Proximal policy optimization algorithms.
\newblock \emph{arXiv preprint arXiv:1707.06347}, 2017.

\bibitem[Soroush~Nasiriany(2019)]{nasiriany2019planning}
Steven Lin Sergey~Levine Soroush~Nasiriany, Vitchyr~Pong.
\newblock Planning with goal-conditioned policies.
\newblock 2019.

\bibitem[Todorov et~al.(2012)Todorov, Erez, and Tassa]{mujoco}
Emanuel Todorov, Tom Erez, and Yuval Tassa.
\newblock Mujoco: A physics engine for model-based control.
\newblock In \emph{IROS 2012}, pp.\  5026--5033. IEEE, 2012.

\end{thebibliography}
\bibliographystyle{iclr2020_conference}

\newpage

\appendix
\section{Appendix}

In this section, we provide more details about our method and baselines. We also describe the experimental setup, and provide a detailed analysis of the results. 

\subsection{Data Augmentation}
\label{data-aug}

We described the intuitions behind data augmentation in Section \ref{fig:goal-conditioned-batch-rl}. Here, we provide the algorithm to collect augmented samples from the environment after the on-policy data collection stage in the goal-conditioned batch RL setting.

\begin{algorithm}
  \caption{Data augmentation}
  \label{alg:data-augmentation}
    \begin{algorithmic}
      \STATE Given $\mathcal{D} = \{\tau_1, \tau_2, ..., \tau_N\}$, environment $e$
      \STATE Initialize $\tilde{\mathcal{D}} = \{\}$
      \FOR{$\tau = \{s_1, a_1, g_1, ..., s_T, a_T, g_T\} \in \mathcal{D}$}
        \STATE Sample $\theta \sim [0, 2\pi)$
        \STATE $\tilde{s}_1 \leftarrow$ $e.rot(s_1, \theta)$
        \STATE Initialize $\tilde{\tau} = \{\}$
        \FOR{$t=1$ {\bfseries to} $T$}
            \STATE $\tilde{a}_t = a_t$
            \STATE $\tilde{s}_{t+1} = e.step(\tilde{a}_t)$
            \STATE Record one-step goal reached $\tilde{g}_t$
            \STATE $\tilde{\tau} \leftarrow \tilde{\tau} \cup \{\tilde{s}_t, \tilde{a}_t, \tilde{g}_t\}$
        \ENDFOR
        \STATE $\tilde{\mathcal{D}} \leftarrow \tilde{\mathcal{D}} \cup \tilde{\tau}$
      \ENDFOR
      \STATE $\mathcal{D}_{aug} = \mathcal{D} \cup \tilde{\mathcal{D}}$
    \end{algorithmic}
\end{algorithm}

\subsection{Learning a Na\"ive Goal-conditioned Policy}
\label{learning-gcp}
Once we have a dataset consisting of $(s, a, g)$ tuples, we use it to train the goal-conditioned policy $\pi_g \colon \mathcal{S} \times \mathcal{G} \rightarrow \mathcal{A}, \pi_g(s, g) = a$. We train a multilayer perceptron (MLP) using the objective $\mathcal{L}_{\pi_g}$.
\begin{equation}\label{eq:pi-g}
    \mathcal{L}_{\pi_g} = \frac{1}{M}\sum_{i=1}^M ||\pi_g(s_i, g_i) - a_i||_2^2
\end{equation}
where $M$ is the number of samples in the dataset. The dataset can be either $\mathcal{D}$ consisting of on-policy samples, or augmented samples $\mathcal{D}_{aug}$, or even randomly collected samples.

\begin{algorithm}
  \caption{Na\"ive goal-conditioned policy}
  \label{alg:gcp}
\begin{algorithmic}
  \STATE Given $\mathcal{D} = \{\tau_1, \tau_2, ..., \tau_N\}$ \\or $\mathcal{D}_{aug} = \{\tau_1, \tilde{\tau}_1, \tau_2, \tilde{\tau}_2, ..., \tau_N, \tilde{\tau}_N\}$, learning rate $\eta$
  \STATE Given policy $\pi_g$ with parameters $\phi_g$
  \STATE Initialize $\phi_g$ randomly
  \REPEAT
      \FOR{$\tau = \{s_1, a_1, g_1, s_2, a_2, g_2, ..., s_T, a_T, g_T\} \in \mathcal{D}$}
        \STATE Compute $\mathcal{L}_{\pi_g}$ using Eq. \ref{eq:pi-g}
        \STATE $\phi_g \leftarrow \phi_g - \eta \nabla \mathcal{L}_{\pi_g}$
      \ENDFOR
  \UNTIL{$n$ iterations}
\end{algorithmic}
\end{algorithm}

\subsection{Our Approach: Enforcing Equivalence}

\begin{figure}
  \begin{center}
  \includegraphics[width=\linewidth]{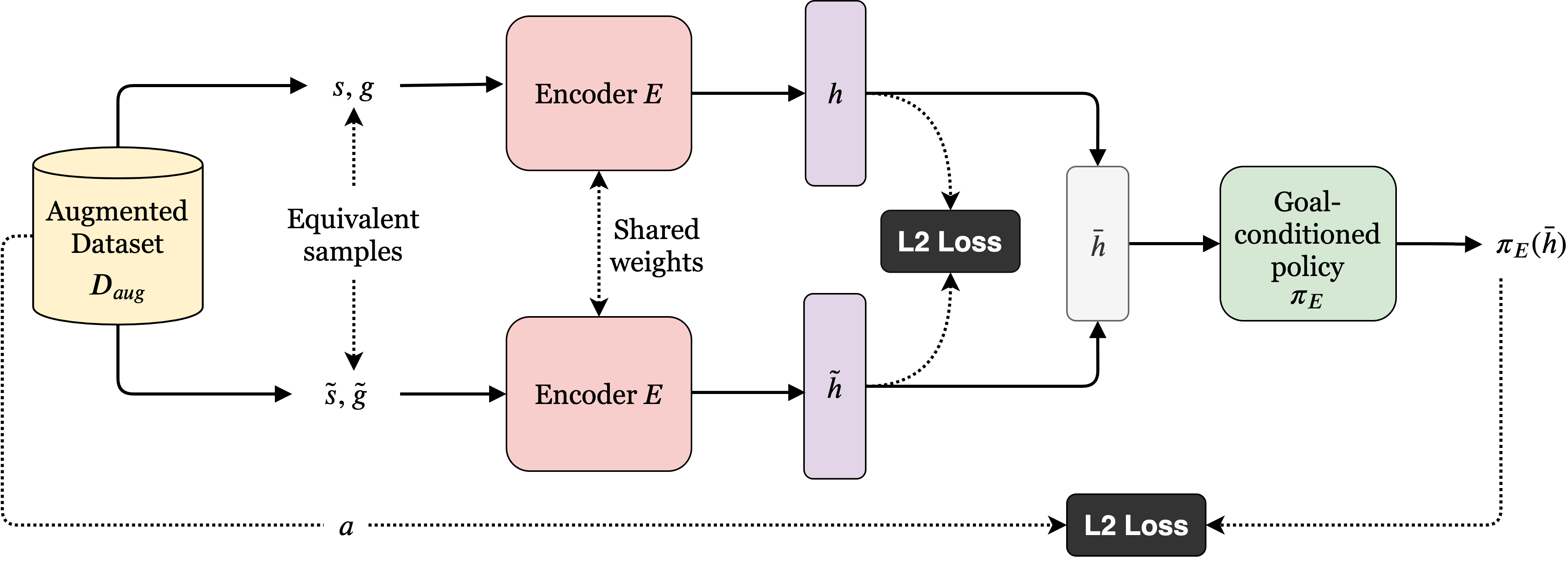}
  \caption{Our approach involves two components: encoder $E$ that generates embeddings $h$ and $\tilde{h}$ for a pair of equivalent samples $(s, g)$ and $(\tilde{s}, \tilde{g})$ respectively using shared weights, and goal-conditioned policy with equivalence $\pi_E$ that takes as input the mean of these embeddings $\bar{h}$ to output the required action.}
  \label{fig:our-approach}
  \end{center}
\end{figure}

The training procedure of our approach that enforces equivalence on augmented samples $(s, g)$ and $(\tilde{s}, \tilde{g})$ is depicted in Fig. \ref{fig:our-approach}.

\subsection{Experimental Details}
In this section, we discuss the details of all the experiments performed in the standard RL, goal-conditioned RL and goal-conditioned batch RL settings. All experiments were trained and tested on an Nvidia GeForce GTX 1080 GPU.

\subsubsection{Environments}
\begin{figure}
  \centering
  \includegraphics[width=0.318\linewidth]{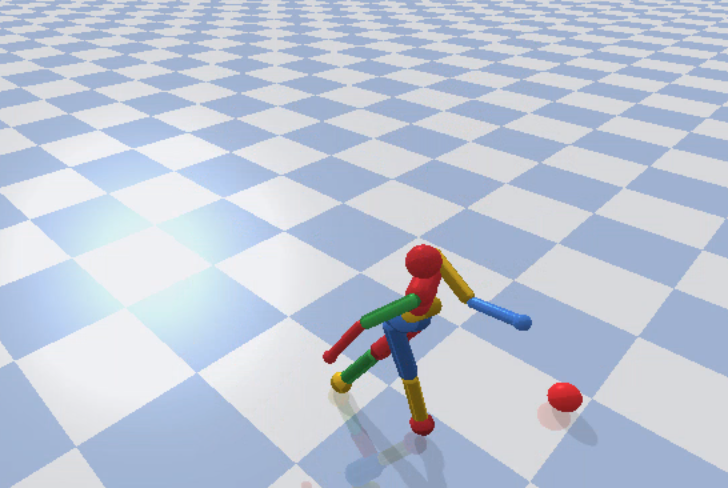}
  \includegraphics[width=0.316\linewidth]{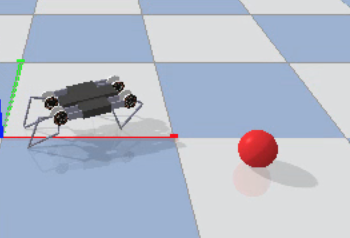}
  \includegraphics[width=0.312\linewidth]{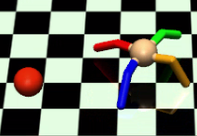}
  \caption{We report the results of our algorithm and baselines on the three locomotion environments shown above: Humanoid, Minitaur, and Ant. The red sphere indicates the goal in each environment.}
  \label{fig:envs}
\end{figure}

We show the results of our approach, compared with other baselines, on the 3-D locomotion environments: Ant (\cite{mujoco, gym}), Minitaur and Humanoid (\cite{pybullet}) (shown in Fig. \ref{fig:envs}). The Humanoid is a bipedal agent with a 43-D state space and 17-D action space. The Ant is a quadrupedal agent with a 111-D state space and 8-D action space. The Minitaur is also a quadrupedal agent with a 17-D state space and 8-D action space.

\subsubsection{Reward Functions}

The reward function for the policy used for data collection in the goal-conditioned batch RL setting includes the control and contact costs, and a reward for moving forward to the right.

The reward function for training a goal-conditioned policy using standard RL includes the control and contact costs, and a reward for moving in the direction of the goal.

In the goal-conditioned RL setting, we consider two scenarios: sparse and dense rewards. In the sparse reward experiments, the agent receives a reward of 0 if it reaches within a certain distance of the goal, and -1 otherwise. In the dense reward scenario, the agent receives a weighted sum of the control and contact costs, and the distance to the goal, as its reward.

\subsubsection{Data Collection}
In this section, we provide details about the data collection process for the goal-conditioned batch RL setting.

The batch data we collect for the goal-conditioned batch RL experiments (discussed in the main paper), consist of three kinds of data: random samples, on-policy samples, and augmented samples. We collect the same number of samples for all baselines, and also train the standard and goal-conditioned RL policies using the same number of environment interactions.

\textbf{Ant.} We use 2 million timesteps of environment interaction for each method. For the standard RL and goal-conditioned RL methods, this means training a policy for 2 million timesteps. The data collection for the goal-conditioned batch RL setting is done in the following manner. For the baseline with random samples, the batch consists of 2 million timesteps of the agent taking random actions in the environment. The baseline with on-policy samples has a batch consisting of 2 million samples collected while training a standard RL (\cite{ppo, pmlr-v80-haarnoja18b}) algorithm for the locomotion task. The baseline with augmented samples, and our final approach that enforces equivalence, have a batch consisting of 1 million samples collected while training the standard RL algorithm, and 1 million samples that are equivalent to those collected at training time. The data collection policy is learnt using PPO (\cite{ppo}).

\textbf{Minitaur.} We use 4 million timesteps of environment interaction. The procedure for training the standard and goal-conditioned RL methods, as well as data collection for the goal-conditioned batch RL methods is the same as that for the Ant (discussed above). The data collection policy is learnt using PPO (\cite{ppo}).

\textbf{Humanoid.} For the Humanoid environment, the first 5 million timesteps of training a standard RL algorithm for locomotion consisted of predominantly bad samples that led to the Humanoid falling down very early in the episode. Thus, instead of using those samples, we froze training after 5 million timesteps, and used the policy trained to collect 10 million samples. The standard and goal-conditioned RL methods were trained for 15 million timesteps, for fair evaluation. The procedure for collecting the batch data is the same as that discussed above for Ant and Minitaur. The data collection policy is learnt using SAC (\cite{pmlr-v80-haarnoja18b}).

\subsubsection{Network architecture}
There are 3 models we learn in the goal-conditioned batch RL methods: the na\"ive goal-conditioned policy $\pi_g$, the encoder $E$, and the goal-conditioned policy with equivalence $\pi_E$. We also have the data collection policy $\pi_{data}$ for batch data. The standard and goal-conditioned RL algorithms also involve learning the policies $\pi_{standard}$ and $\pi_{goal}$ respectively. All these networks are MLPs with 2 hidden layers and $tanh$ activation.

\begin{table}
    \centering
    \begin{tabular}{|l|l|l|}
        \hline
        \textbf{Environment} & \textbf{Model} & \textbf{Architecture}\\
        \hline
        \multirow{6}{*}{Humanoid} &
            $\pi_{data}$ & $256 \times 256$\\
            \cline{2-3}
            & $\pi_{standard}$ & $256 \times 256$\\
            \cline{2-3}
            & $\pi_{goal}$ & $64 \times 64$\\
            \cline{2-3}
            & $\pi_g$ & $256 \times 256$\\
            \cline{2-3}
            & $E$ & $256 \times 256$\\
            \cline{2-3}
            & $\pi_E$ & $256 \times 256$\\
            \hline
        \multirow{6}{*}{Minitaur} &
            $\pi_{data}$ & $256 \times 256$\\
            \cline{2-3}
            & $\pi_{standard}$ & $256 \times 256$\\
            \cline{2-3}
            & $\pi_{goal}$ & $64 \times 64$\\
            \cline{2-3}
            & $\pi_g$ & $256 \times 256$\\
            \cline{2-3}
            & $E$ & $256 \times 256$\\
            \cline{2-3}
            & $\pi_E$ & $50 \times 50$\\
            \hline
        \multirow{6}{*}{Ant} &
            $\pi_{data}$ & $64 \times 64$\\
            \cline{2-3}
            & $\pi_{standard}$ & $64 \times 64$\\
            \cline{2-3}
            & $\pi_{goal}$ & $64 \times 64$\\
            \cline{2-3}
            & $\pi_g$ & $256 \times 256$\\
            \cline{2-3}
            & $E$ & $256 \times 256$\\
            \cline{2-3}
            & $\pi_E$ & $50 \times 50$\\
            \hline
            
    \end{tabular}
    \caption{Training details for Humanoid, Minitaur, Ant. Note that the hyperparameters for the standard and goal-conditioned RL experiments, as well as those for the policy used for data collection (\cite{ppo, pmlr-v80-haarnoja18b, andrychowicz2017hindsight, ddpg}), have been taken from existing implementations of these algorithms (\cite{baselines, stable-baselines}).}
    \label{tab:training}
\end{table}

\subsubsection{Hyperparameters}

For the standard RL and goal-conditioned RL baselines we compare our approach with, we use the hyperparameters provided in existing implementations of these algorithms (\cite{baselines, stable-baselines}). The hyperparameters involved in training all the goal-conditioned batch RL methods are:

\textbf{Weighted loss parameter $\lambda$.} The loss function for our approach is a weighted sum of the encoder loss $L_{enc}$ and the policy loss $L_{\pi_E}$. We set $\lambda = 0.25$ for all environments.

\textbf{Encoder output dimension $k$.} The output of the encoder in our approach is a $k$-dimensional vector. We set $k = 10$ for the Ant and Minitaur environments, and $k = 50$ for the Humanoid environment.

\textbf{Optimizer.} We use the Adam optimizer (\cite{DBLP:journals/corr/KingmaB14}) with a learning rate $0.001$ and a batch size of $512$ for all goal-conditioned batch RL methods across all environments.

\subsubsection{Baselines}

In the standard RL baseline, we used SAC (\cite{pmlr-v80-haarnoja18b}) for the Humanoid and PPO (\cite{ppo}) for the Ant and Minitaur to train the policy.

In the goal-conditioned RL baseline with sparse rewards, we used SAC (\cite{pmlr-v80-haarnoja18b}) as the off-policy RL algorithm with HER (\cite{andrychowicz2017hindsight}) for all 3 environments. In the dense reward setting, we used SAC for the Humanoid and Minitaur, and DDPG (\cite{ddpg}) for the Ant.

\subsubsection{Test setup}
At test time, the agent is rotated by a random angle. The target is set at a distance of $\sim2-5$ units from the agent at an angle of $[-45^{\circ}, 45^{\circ}]$ to the agent's orientation in the case of the Ant and Humanoid environments. For Minitaur, we set the target at a distance of $\sim1.5-2.5$ units from the agent at an angle of $[-45^{\circ}, 45^{\circ}]$ to the agent's orientation. At each point, we replace the actual goal with the unit vector in the direction of the goal as the input.

Each episode consists of a maximum of 1000 steps for each environment, and the episode terminates when the agent reaches the goal, or falls down/dies. We report the closest distance from the target that the agent is able to reach, for each episode. We select 10 random seeds and test the performance of each method on 1000 episodes for each random seed. In order to ensure that the comparison between all methods is indeed fair, we set the initial configuration of the agent and the target to be the same across all methods at test time.

\subsection{Analysis of Results}

\textbf{Why does random sampling in goal-conditioned batch RL show the worst performance?}

Using random samples as the batch data in the goal-conditioned batch RL setting leads to the worst results in all environments. This happens because all agents we consider are high-dimensional continuous control agents, which cannot be expected to learn good goal-conditioned policies from random data. 

\begin{figure*}
  \centering
  \includegraphics[width=0.47\linewidth]{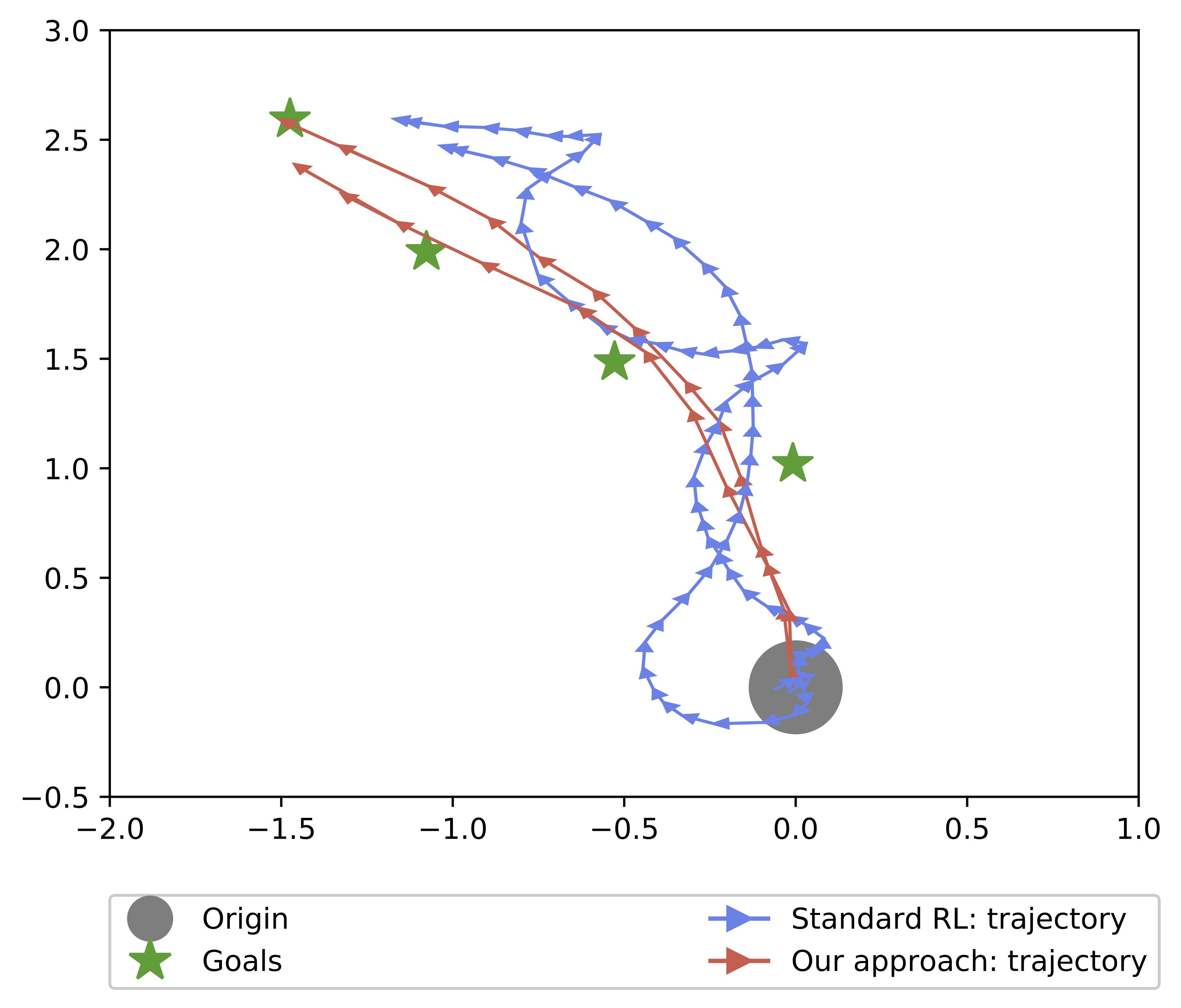}
  \includegraphics[width=0.45\linewidth]{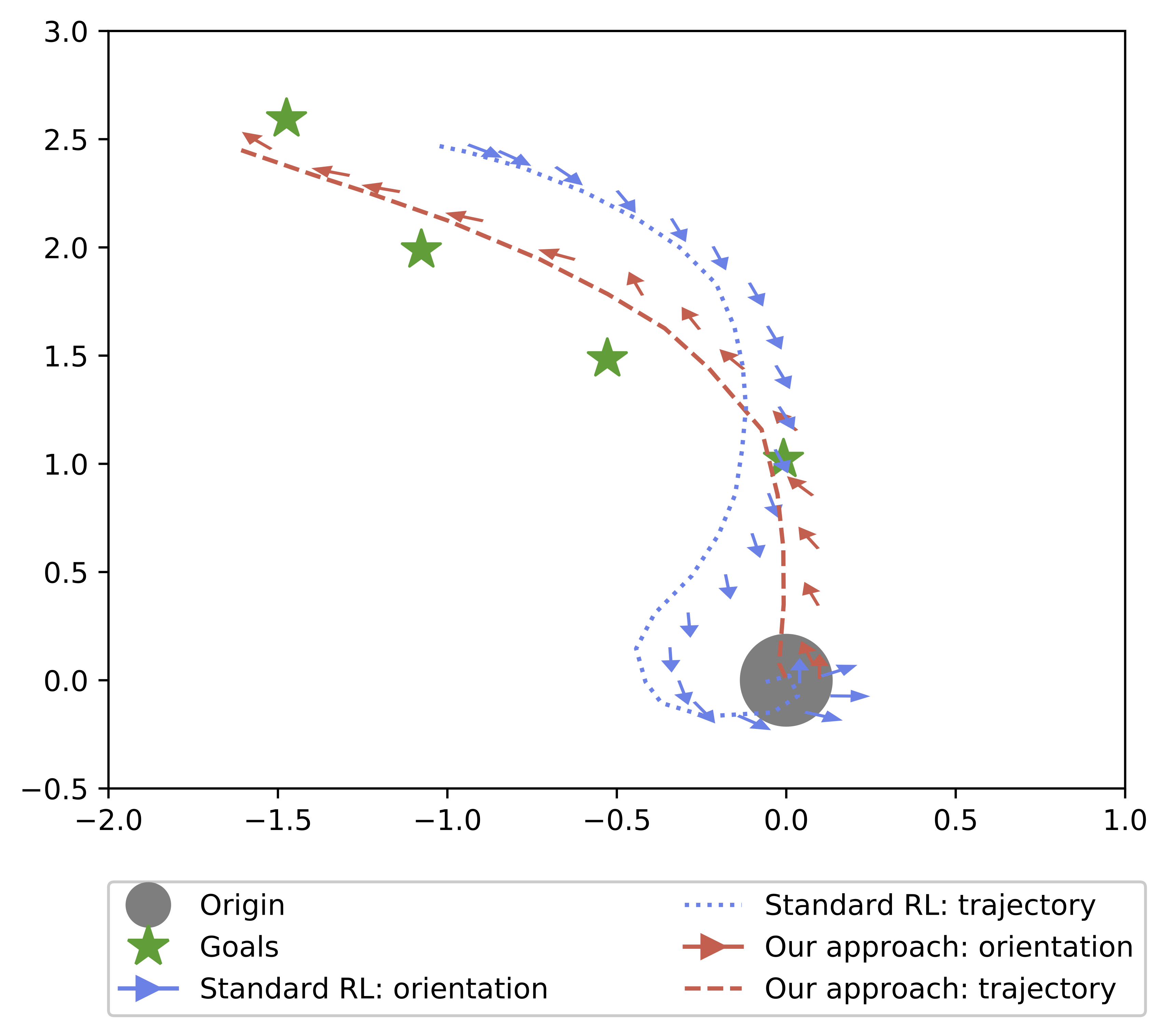}
  \caption{We compare the performance of the best standard RL baseline with our approach qualitatively. The first plot shows two successful trajectories through 4 consecutive goals. The second plot shows the orientation of the agent through one successful trajectory for both methods. The quality of the standard RL baseline lower than that of our approach in terms of the trajectory followed, orientation of the agent, and the speed of locomotion.}
  \label{fig:qualitative}
\end{figure*}

\textbf{Does on-policy data improve performance over random data in the goal-conditioned batch RL setting?}

Learning from a batch that \textit{does} contain on-policy samples, i.e.\ samples collected while training a locomotion policy in a single direction, does improve performance over random samples, but it is unfortunately unable to exceed the performance of the standard RL and goal-conditioned RL methods in all environments due to poor generalization to goals that lie out of its training distribution. However, for goals that lie within its training distribution, it exhibits smooth goal-directed locomotion, as opposed to the poor performance of goal-conditioned batch RL with random samples. This is evident from the violin plots shown in Fig. \ref{fig:violin_plots}: the plot for goal-conditioned batch RL with on-policy samples has a small peak away from the mean, which consists of goals belonging to its training distribution.

\textbf{How does data augmentation affect performance?}

Learning from augmented samples in the goal-conditioned batch RL setting improves performance over standard and goal-conditioned RL baselines, as well as over the two preliminary goal-conditioned batch RL baselines. While the gain in performance over the two simpler goal-conditioned batch RL baselines is for obvious reasons, this method works better than even standard RL and goal-conditioned RL because it consists of samples that take consistent actions to reach the goal in the case of \textit{equivalent} state-goal configuration. In other words, this means that if we have two trajectories in the augmented dataset $\tau = \{s_1, a_1, g_1, s_2, a_2, g_2, ..., s_T, a_T, g_T\}$ and $\tilde{\tau} = \{\tilde{s}_1, \tilde{a}_1, \tilde{g}_1, \tilde{s}_2, \tilde{a}_2, \tilde{g}_2, ..., \tilde{s}_T, \tilde{a}_T, \tilde{g}_T\}$, where $(s_1, g_1) \sim (\tilde{s}_1, \tilde{g}_1)$, the augmented samples collected would have $a_1 = \tilde{a}_1, a_2 = \tilde{a}_2, ..., a_T = \tilde{a}_T$, but the same policy may not be learnt by standard RL and goal-conditioned RL methods.

\textbf{Why does incorporating equivalence into the learning algorithm improve performance over using augmented samples?}

While augmenting the dataset with equivalent trajectories already beats all other baselines in terms of performance, enforcing equivalence between equivalent trajectories is able to achieve a far higher generalization to goals not observed in the training distribution. This is because the encoder learns the same embeddings for equivalent state-goal configurations. Thus, for any goal $g$ that lies outside the training distribution, the agent starting from state $s$ will be able to achieve it successfully if $(\tilde{s}, \tilde{g})$, where $(s, g) \sim (\tilde{s}, \tilde{g})$, lies in its training distribution.

\textbf{Why does standard RL perform better than goal-conditioned RL methods?}

An interesting observation is that standard RL works the same as or, in some cases, better than goal-conditioned RL for the goal-directed locomotion task in the sparse-reward, as well as the dense-reward settings. While this was initially surprising, we realized that since goal-directed locomotion is a difficult task, it would take a very large number of samples for the agent to reach a goal and receive a positive reward. Thus, we observe that in each case, the agent trained with dense reward (i.e.\ including contact and control costs in the reward function) performs better than that trained with sparse reward.

\textbf{Is there any qualitative difference between the policy learnt using our approach v/s standard RL?}

We perform an additional experiment to check the quality of trajectories that the agent follows when it is trained using our approach v/s standard RL, which is the best baseline. In this experiment, we analyze \textit{longer} trajectories for better qualitative comparison. We do this by setting a new goal as soon as the agent achieves one goal, instead of starting a new episode. We consider a goal to be achieved if the agent is within a certain distance from the goal (here, the distance is 0.5 units). We plot the results of two successful trajectories of standard RL and our approach on 4 successive goals, and show the results in Fig. \ref{fig:qualitative}. The first plot shows the path followed by the agent while it tries to achieve these goals, and the second plot shows the direction in which the agent is facing while moving in the direction of the goal i.e.\ its orientation. It is evident from the first plot that our approach follows a direct path to reach each goal, and it does so much faster than standard RL, which takes a much higher number of steps to reach all goals. Furthermore, if we see the second plot, the agent trained using our approach always faces the direction of locomotion, whereas the agent trained using standard RL methods faces the direction opposite to that of locomotion i.e.\ it walks backward. This clearly reflects the qualitative superiority of our method, since the agent learns a representation of equivalent states and goals, and thus, takes uniform actions in all directions, as opposed to an agent trained using standard RL methods.

\begin{figure*}
  \centering
  \includegraphics[width=0.43\linewidth]{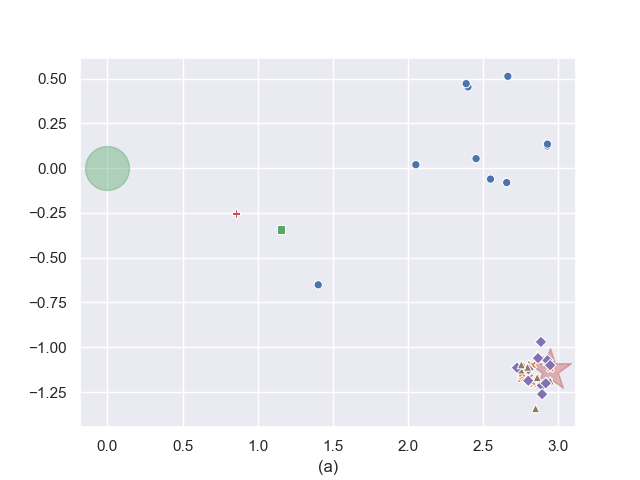}
  \includegraphics[width=0.43\linewidth]{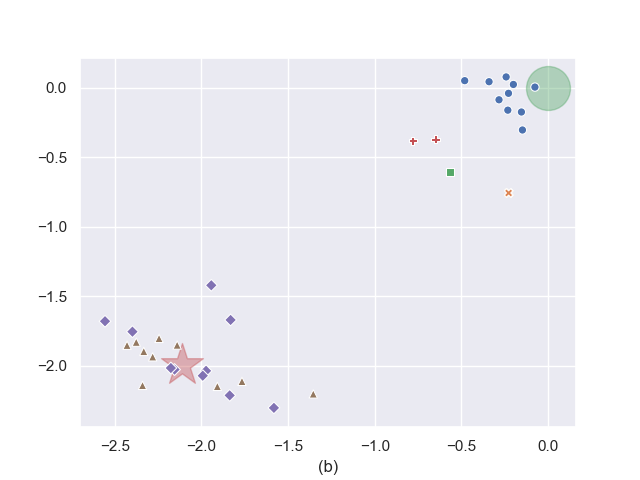}\\
  \includegraphics[width=0.85\linewidth]{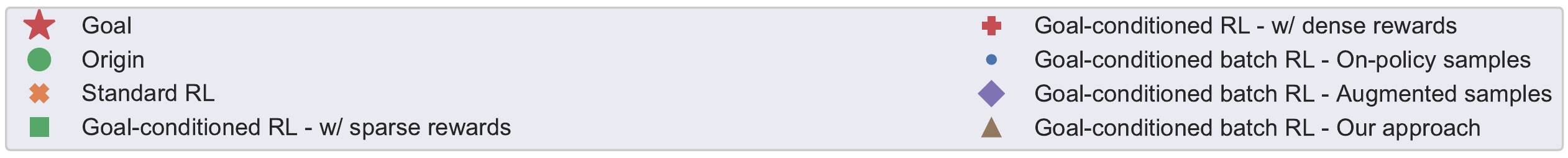}
  \caption{Qualitative comparison between different algorithms at test time for the Humanoid.}
  \label{fig:qual-humanoid}
\end{figure*}

\begin{figure*}
  \centering
  \includegraphics[width=0.43\linewidth]{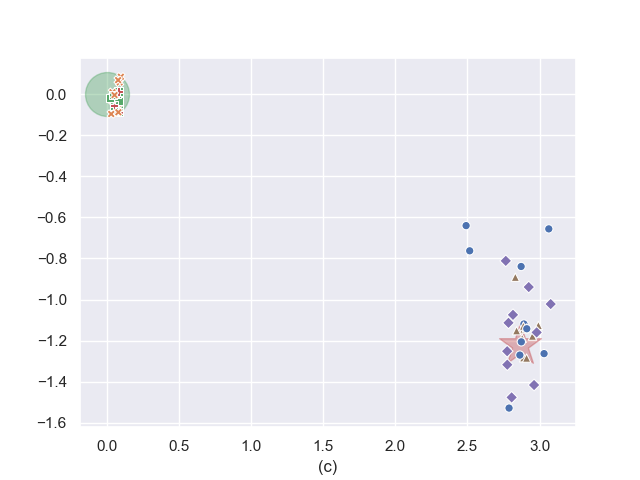}
  \includegraphics[width=0.43\linewidth]{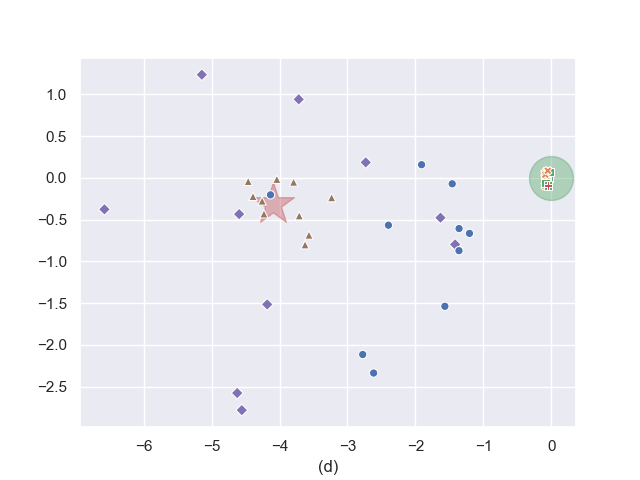}\\
  \includegraphics[width=0.85\linewidth]{plots/legend.pdf}
  \caption{Qualitative comparison between different algorithms at test time for the Ant.}
  \label{fig:qual-ant}
\end{figure*}

\subsection{Additional Analysis of Results}

The violin plots in the main paper show a slight peak away from the mean for a number of baselines. This peak is a result of the agent being able to reach goals that lie within its training distribution. We conduct an additional experiment to provide qualitative proof for this. We take one instance each of a bipedal and a quadrupedal agent, the Humanoid and the Ant, and show results in Fig. \ref{fig:qual-humanoid} and Fig. \ref{fig:qual-ant} respectively. We fix the initial state of the agent and generate goals in a specific region. We plot the results of 10 best episodes for each method for the Humanoid and Ant environments. The standard RL baseline is sometimes successful in reaching the goal. The goal-conditioned RL algorithm (HER) with sparse or dense rewards fails early in the episode, because HER works better in sparse reward scenarios, and it takes a large number of samples to learn a good locomotion policy with only sparse rewards. In the goal-conditioned batch RL setting, the baseline trained using only on-policy samples can reach goals lying within its training distribution, on the right, but fails on other goals. Both our proposed ideas: data augmentation and enforcing equivalence, result in trajectories that reach closest to the goal in all cases.

\end{document}